\documentclass[twoside]{article}

\usepackage[accepted]{aistats2020}

\usepackage[utf8]{inputenc} %
\usepackage[T1]{fontenc}    %
\usepackage{hyperref}       %
\usepackage{url}            %
\usepackage{booktabs}       %
\usepackage{amsfonts}       %
\usepackage{nicefrac}       %
\usepackage{microtype}      %

\usepackage{graphicx}
\usepackage{natbib}
\usepackage{subfig}
\usepackage{dblfloatfix} %
\usepackage{float}
\usepackage{bm}
\usepackage{amsmath}
\usepackage{amssymb}
\usepackage{todonotes}
\usepackage{url}
\usepackage{multirow}
\usepackage{appendix}
\usepackage[makeroom]{cancel}
\usepackage{paralist}
\usepackage{hyperref}
\usepackage{wrapfig}

\usepackage{cleveref}
\crefname{appsec}{Appendix}{Appendices}

\newcommand{\GP}{\ensuremath{\mathcal{GP}}}
\newcommand*\diff{\mathop{}\!\mathrm{d}}

\DeclareMathOperator*{\given}{|}
\DeclareMathOperator{\E}{\mathbb{E}} %
\DeclareMathOperator{\cov}{\operatorname{Cov}} %
\renewcommand{\Re}{\mathbb{R}} %
\newcommand{\KL}{\operatorname{KL}}
\DeclareMathOperator{\Gauss}{\mathcal{N}} %

\renewcommand{\vec}[1]{\bm{\mathrm{#1}}} %
\newcommand{\mat}[1]{\bm{\mathrm{#1}}} %
\newcommand{\BigO}{\mathcal{O}} %

\usepackage{xspace}

\newcommand{\MZ}{\mat{Z}}

\newcommand{\MS}{\mat{S}}

\newcommand{\MK}{\mat{K}}
\newcommand{\Eye}{\mat{I}}

\newcommand{\Kxx}{\MK_{\textsc{X}\textsc{X}}}

\newcommand{\vx}{{\vec{x}}}
\newcommand{\vu}{{\vec{u}}}
\newcommand{\vf}{{\vec{f}}}

\newcommand{\vy}{{\vec{y}}}
\newcommand{\vm}{{\vec{m}}}
\newcommand{\vz}{{\vec{z}}}
\newcommand{\vk}{{\vec{k}}}

\newcommand{\vh}{{\vec{h}}}

\let\originalleft\left
\let\originalright\right
\renewcommand{\left}{\mathopen{}\mathclose\bgroup\originalleft}
\renewcommand{\right}{\aftergroup\egroup\originalright}

\begin{document}

\runningtitle{Bayesian Image Classification with Deep Convolutional Gaussian Processes}

\twocolumn[

\aistatstitle{Bayesian Image Classification with\\ Deep Convolutional Gaussian Processes}

\aistatsauthor{ Vincent Dutordoir \And Mark van der Wilk \And  Artem Artemev \And James Hensman }

\aistatsaddress{PROWLER.io, Cambridge, United Kingdom} ]

\begin{abstract}
In decision-making systems, it is important to have classifiers that have calibrated uncertainties, with an optimisation objective that can be used for automated model selection and training. Gaussian processes (GPs) provide uncertainty estimates and a marginal likelihood objective, but their weak inductive biases lead to inferior accuracy. This has limited their applicability in certain tasks (e.g.~image classification). We propose a translation-insensitive convolutional kernel, which relaxes the translation invariance constraint imposed by previous convolutional GPs. We show how we can use the marginal likelihood to learn the degree of insensitivity. We also reformulate GP image-to-image convolutional mappings as multi-output GPs, leading to deep convolutional GPs. We show experimentally that our new kernel improves performance in both single-layer and deep models. We also demonstrate that our fully Bayesian approach improves on dropout-based Bayesian deep learning methods in terms of uncertainty and marginal likelihood estimates.
\end{abstract}

\section{INTRODUCTION}

To be useful in the real world, decision-making systems have to be able to represent uncertainty. This enables the system to gracefully deal with unseen or special cases and, for example, hand over control to a human operator when the uncertainty is high. It is also crucial to have an accurate measure of uncertainty when making automated decisions based on machine classification (e.g.~medical diagnosing).

Recently, Bayesian deep learning methods based on dropout have been empirically successful in improving the robustness of Deep Neural Nets (DNN) predictions \citep{gal2016dropout}, but it is unclear to what extent they accurately approximate the true posteriors \citep{hron2018dropout}. They also do not deliver on an important promise of the Bayesian framework: automatic regularisation of model complexity which allows the training of hyperparameters \citep{rasmussen2001occam}. Current marginal likelihood estimates are not usable for hyperparameter selection, and the strong relationship between their quality, and the quality of posterior approximations suggests that further improvements are possible with better Bayesian approximations.

We are interested in Gaussian processes (GPs) as an alternative building block for creating deep learning models with the benefits of Bayesian inference. 
Their practical application has been limited due to their large computational requirements for big datasets, and due to the limited inductive biases that they can encode. In recent years, however, advances in stochastic variational inference have enabled GPs to be scaled to large datasets for both regression and classification models \citep{hensman2013,hensman2015scalable}. More sophisticated model structures that are common in the deep learning community, such as depth \citep{damianou2013deep} and convolutions \citep{van2017convolutional}, have been incorporated as well. Notably, inference is still accurate enough to provide marginal likelihood estimates that can be used for hyperparameter selection (e.g.~\citep{vdw2018invariances}).

In this work, we focus on creating models for image inputs. While existing GP models with kernels like the Squared Exponential (SE) kernel have the capacity to learn any well-behaved function when given infinite data \citep[chapter~7]{rasmussen2006}, they are unlikely to work well for image tasks with realistic dataset sizes. Local kernels, like the SE, constrain only functions in the prior to be smooth, and allow the function to vary along any direction in the input space. This will allow these models to generalise only in neighbourhoods near training data, with large uncertainties being predicted elsewhere. This excessive flexibility is a particular problem for images, which have high input dimensionality, while exhibiting a large amount of structure. When designing Bayesian models it is crucial to think about sensible inductive biases to incorporate into the model. For instance, convolutional structure has been widely used to address this issue \citep{lecun1989backpropagation,goodfellow2016deep}. \Citet{van2017convolutional} introduced this structure into a single-layer GP, together with an efficient inference scheme, and showed that this improved performance on image classification tasks. Recently, \citet{blomqvist2019deep} added convolutional structure to deep GPs, which led to deep convolutional Gaussian processes (DCGPs).

\paragraph{Contributions} We start by re-formulating the hidden layers of a DCGP as a correlated multi-output GP. This is a convenient abstraction that enables us to code the convolutional layers in our efficient multi-output GP framework \citep{gpflow2}. We then identify that translational invariant properties of current convolutional models are too restrictive and limits performance. To remedy this, we introduce the Translation Insensitive Convolutional Kernel (TICK), which removes the restriction of requiring identical outputs for identical patch inputs. 
We compare our model to current convolutional GPs, and find improvement in performance in both accuracy and uncertainty quantification. 
Comparing our model to dropout-based Bayesian deep learning methods, we show how our model is competitive in terms of accuracy but also comes with the desirable properties of a truly Bayesian model: a marginal likelihood for model selection, automatically tuning of hyperparameters, and calibration.

\section{BACKGROUND}

\paragraph{Gaussian Process Models}
\label{sec:gp}

Gaussian processes (GPs) \citep{rasmussen2006} are non-parametric distributions over functions similar to Bayesian neural networks. The core difference is that neural networks represent distributions over functions through distributions on weights, while a Gaussian process specifies a distribution on function values at a collection of input locations. Using this representation allows us to use an infinite number of basis functions, while still enables Bayesian inference \citep{neal1996bayesian}. In a GP, the joint distribution of these function values is Gaussian and is fully determined by its mean $\mu(\cdot)$ and covariance (kernel) function $k(\cdot, \cdot)$. Taking the mean function to be zero without loss of generality, function values at inputs $X = \left\{\vx_m\right\}_{m=1}^M$ are distributed as $f(X) \sim \Gauss\left(f(X); \mu(X), \Kxx\right)$, where $\left[\Kxx\right]_{ij} = k(\vx_i, \vx_j)$. The Gaussianity, and the fact that we can manipulate function values at some finite points of interest without taking the behaviour at any other points into account (the marginalisation property) make GPs particularly convenient to manipulate and use as priors over functions in Bayesian models.

\paragraph{Convolutional Gaussian Processes}
\label{sec:cgp}

\Citet{van2017convolutional} construct the convolutional kernel for functions from images of size $D = W\!\times\! H$ to real-valued responses $f\!:\! \Re^D \!\to\! \Re$. Their starting point is a \emph{patch response function} $g\!:\! \Re^{E} \!\to\! \Re$ operating on patches of the input image of size $E = w\!\times\! h$. The output for a particular image is found by taking a sum of the patch response function applied to all patches of the image. A vectorised image $\vx$ of height $H$ and width $W$ contains $P = (H - h + 1) \times (W - w + 1)$ overlapping patches when we slide the window one pixel at a time (i.e.~a vertical and horizontal stride of 1). We denote the $p^{\text{th}}$ patch of an image as $\vx^{[p]}$. Placing a GP prior on $g(\cdot) \sim \GP(0, k_g(\cdot, \cdot))$ implies:
\begin{equation}
\begin{gathered}
    \label{eq:conv-gp}
    f(\vx) = \sum_{p=1}^P g\left(\vx^{[p]}\right)\\ \implies f(\vx) \sim \GP\left(0,  \sum_{p=1}^P \sum_{p'=1}^P k_g\big(\vx^{[p]}, \vx^{[p']}\big)\right).
\end{gathered}
\end{equation}
The convolution kernel places much stronger constraints on the functions in the prior, based on the idea that similar patches contribute similarly to the function's output, regardless of their position. This prior places more mass in functions that are sensible for images, and therefore allow the model to generalise stronger and with less uncertainty than, for example, the SE kernel.
If these assumptions are appropriate for a given dataset, this leads to a model with a higher marginal likelihood and better generalisation on unseen test data.

\paragraph{Deep Gaussian Processes}
Convolutional structure is an example of how the kernel and its associated feature representation influence the performance of a model.
Deep learning models partially automate this feature selection by learning feature hierarchies from the training data. %
The first layers usually identify edges, corners, and other local features, while combining them into more complicated silhouettes further into the hierarchy. Eventually a simple regressor solves the task.

Deep GPs (DGPs) share this compositional nature, by composing layers of GPs \citep{damianou2013deep}. %
They can be defined as $f(\cdot) = f_{L}(\ldots f_2(f_1(\cdot)))$, where each component is a GP, itself $f_{\ell}(\cdot) \sim \GP\left(0, k_{\ell}(\cdot, \cdot)\right)$.
DGPs enable us to specify priors on flexible functions with compositional structure, and open the door to non-parametric Bayesian feature learning. \citet{salimbeni2017doubly} showed that this is crucial to achieve state-of-the-art performance on many datasets, and that DGP models never perform worse than single-layer GPs.

\section{BAYESIAN MODELLING OF IMAGES}

\begin{figure}
	\begin{minipage}[c]{0.48\textwidth}
	\includegraphics[width=0.48\linewidth]{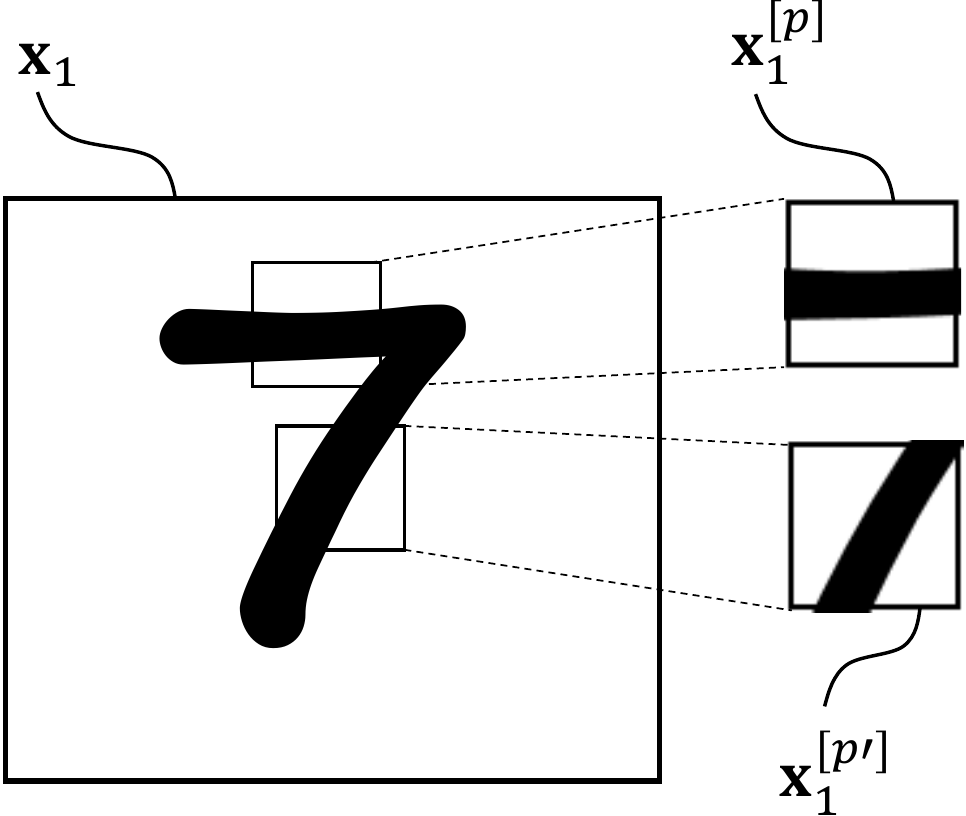}
	\includegraphics[width=0.48\linewidth]{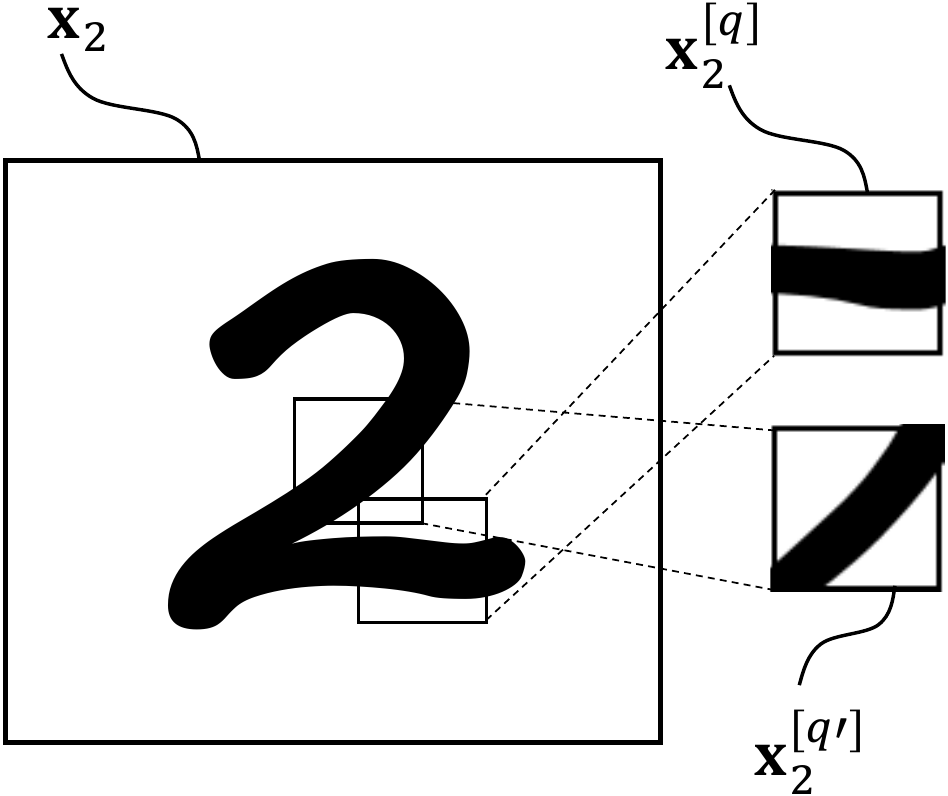}
	\end{minipage}\hfill
	\vspace{.3cm}
	\begin{minipage}[c]{0.5\textwidth}
	\caption{Illustration of why translation \emph{invariance} may be an unrealistic modelling assumption. The highlighted patches are not useful for a translation invariant patch response function $g(\cdot)$, as used in the original convolutional GP \citep{van2017convolutional}, because they appear in both images: only when their relative locations are taken into consideration are these patches useful for classification.
}
	\label{fig:invariance_example}
	\end{minipage}
\end{figure}

\subsection{Limits of the Conv-GP Kernel}

In this section we focus on analysing the behaviour of single-layer convolutional GPs (Conv-GPs), so we can develop improvements in a targeted way.
The convolutional structure in \cref{eq:conv-gp} introduces a form of translation invariance, because the same GP $g(\cdot)$ is used for all patches in the image, regardless of location.
As stated by \citet{liu2018}, a strict form of invariance might or might not be beneficial for certain tasks. For example, in MNIST classification, a horizontal stroke near the top of the digit indicates a `7', while the same stroke near the bottom indicates a `2', as shown in \cref{fig:invariance_example}. The construction of \cref{eq:conv-gp} will apply the same $g(\cdot)$ to each patch in the image, which is undesirable if we wish to distinguish between the two classes by summing $g(\cdot)$'s output.
This also means that it is possible to conceive a complete rearrangement of the image, which appears very different to a human, but is indistinguishable from the original to the convolutional kernel.

\Citet{van2017convolutional} circumvented the translation invariance problem of the Conv-GP by effectively adding a second, linear layer.
By the introduction of weights it is possible to rescale the contribution of each patch, turning the uniform sum of \cref{eq:conv-gp} into a weighted sum $f(\vx) = \sum \nolimits_{p} w_p\,g(\vx^{[p]})$. This is a rudimentary approach which might be both too flexible, (in that it allows wildly varying weights for neighbouring pixels) and not flexible enough, (in that an image evaluation will always be a linear combination of evaluations of $g(\cdot)$ at the input patches).

\begin{figure*}[t]
\centering
\subfloat[][Conv-GP]{
  \includegraphics[width=0.48\linewidth]{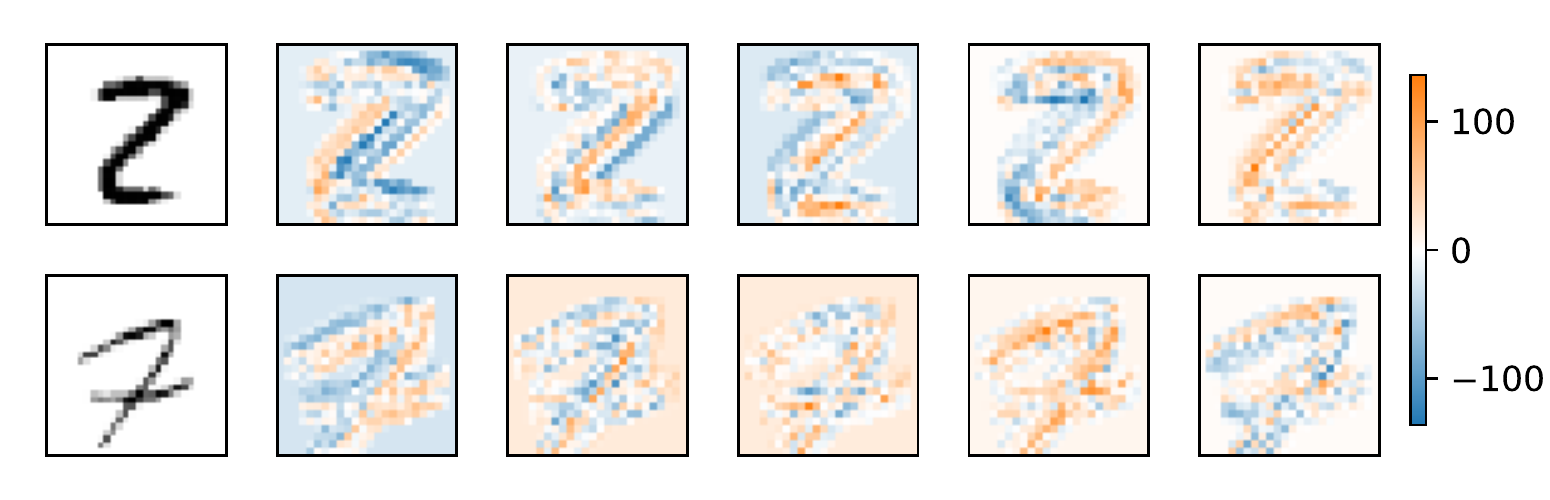}
 }
 \hfill
 \subfloat[][TICK-GP]{
  \includegraphics[width=0.48\linewidth]{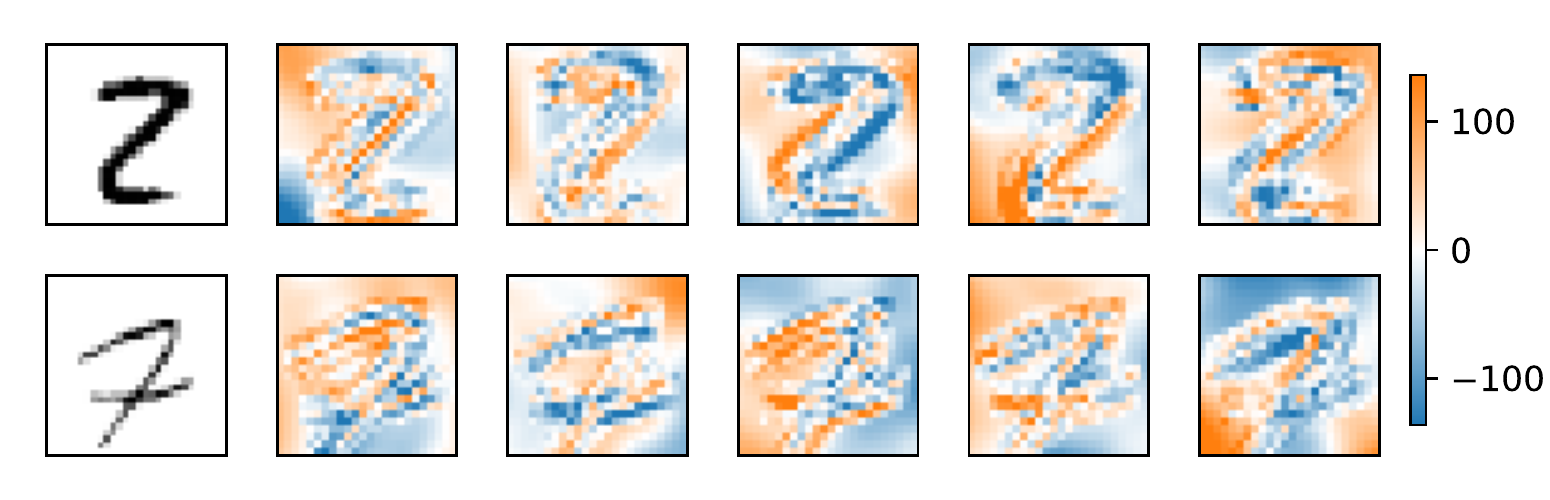}
 }
\caption{We show five samples from the patch response function $g(\cdot)$ after training on MNIST 2 vs 7. The two black-and-white images (left) are the inputs. They were incorrectly classified by the Conv-GP (a), but correctly classified by the TICK-GP (b). The samples show that the posterior of the Conv-GP is overconstrained, noticeable by the paler colours and the even background (see text). \label{fig:mnist_27}}
\end{figure*}

We illustrate the problem of the original Conv-GP being too constrained in \cref{fig:mnist_27}. We trained a model to classify MNIST 2 vs 7 only, and display the deviations from the mean of samples from the posterior of the patch response function $g(\cdot)$. %
On the left (a) we show posterior samples for the original Conv-GP; we show on the right (b) samples from our TICK-GP. Note that all samples in (a) and (b) are plotted using the same colour range. We immediately notice that the samples in (a) are less vibrant than in (b), indicating the smaller variance of the Conv-GP.
The small variance is the result of the Conv-GP being too constrained, which leads to a collapsed posterior that cannot to accommodate for patches that can be both positive and negative (i.e.~those that belong to both classes). We also notice that all background pixels within an image have the exact same value. We discuss the behaviour of the TICK-GP samples in the next section.

\subsection{Translation Insensitive Convolutional Kernel (TICK)}

A better modelling assumption would be to relax the ``same patch, same output'' constraint and have a patch response function $g(\cdot)$ that can vary its output depending on both the patch input and the patch location. We call this property translation \emph{in}sensitivity, and propose a product kernel between the patches and their locations:
\begin{multline}
    \label{eq:tick}
k_g  \big((\vx^{[p]}, p), (\vx^{[p']}, p')\big)  = \\ k_{\text{patch}}\left(\vx^{[p]},\vx^{[p']}\right)\ \times\ 
    k_\text{loc}\left(\ell(p), \ell(p')\right),
\end{multline}
where $\ell(p)$ returns the location of the upper-left corner of the patch in the image, and $k_{\text{patch}}$ and $k_\text{loc}$ are the kernels we use over the patches and patch locations, respectively. We refer to this kernel as the Translation Insensitive Convolutional Kernel (TICK). The term ``insensitive'' was used by \Citet{vdw2018invariances} as a relaxation of invariance. We use the term to indicate that the output is slightly sensitive to translations.

Similar approaches have been suggested in the CNN literature \citep{ghafoorian2017location}, but have not been adopted in popular, recent architectures (e.g. Inception \citep{inception} and DenseNet \citep{densenet}).
A explanation for this is that this parametric approach in neural nets adds a lot more parameters, in the order of $\mathcal{O}(w\,h\,c_{\text{in}}\,c_{\text{out}})$, leading to models that are prone to overfit in the absence of large datasets. 

In TICK we introduce a single hyperparameter, the lengthscale of $k_\text{loc}$, to control only the degree of insensitivity (i.e. the degree to which the output of $g(\cdot)$ depends on the location of the input patch). We will learn this lengthscale and other hyperparameters automatically, using the marginal likelihood. We use \citet[Theorem 4.1.1.]{adler198} to get an intuition in how this parameter effects $g(\cdot)$ for the same patch input depending on its location. If we assume $N_u$ to be the number of times a GP-draw from a stationary kernel $k$ crosses the level $u$ in the unit interval, then
\begin{equation*}
    \E_{g(\cdot)}[N_u] =  \frac{1}{2 \pi} \sqrt{\frac{-k^{''}(0)}{k(0)}} \exp\left(\frac{-u^2}{2~k(0)}\right).
\end{equation*}
A Squared Exponential (SE) kernel for $k_{\text{loc}}(r) = \sigma^2 \exp(- r^2/ \ell^2)$ gives an expected number of zero-crossings $\E[N_0] = (\pi\ell)^{-1}$. 

You can observe that property most easily in \cref{fig:mnist_27} (b), where the lengthscale of the SE in the trained TICK-GP approximated $(\pi/2)^{-1}$, corresponding to $\approx 2$ zero-crossings in the image. 
Inspecting the identical background patches away from the digit, we see that $g(\cdot)$ varies smoothly, and changes sign (i.e. predicts a different class) depending on where background patches are appearing. The mapping of similar patches also varies smoothly across the stroke: the response of horizontal and vertical lines in the image gives only locally similar responses.
We also notice that the samples from the TICK-GP have much larger deviations from the mean, showing that the patch-response function is less constrained and can represent epistemic uncertainty for observing certain patches at certain locations.

\subsection{Deep Convolutional Gaussian Processes}
\label{sec:dcgp}
With the ideas of improved convolutional kernels and deep Gaussian processes in place, it is straightforward to conceive of a model that does both: a deep GP with convolutional kernels at each layer. To do this we need to make these convolutional layers map from images to images, which we do using a multi-output kernel.

We propose a reformulation to the convolutional kernel of \cref{eq:conv-gp}: instead of summing over the patches, we apply $g(\cdot)$ to all patches in the input image. As a result, we obtain a vector-valued function $f: \Re^D \to \Re^P$ defined as
\begin{equation}
    \vf(\vx) = \left\{ f_p(\vx) \right\}_{p=1}^P = \left\{g(\vx^{[p]})\right\}_{p=1}^P \label{eq:deconv-f-def} \,,
\end{equation}
where $f_p(\cdot)$ indicates the $p^{\text{th}}$ output of $f(\cdot)$.
Because the \emph{same} $g(\cdot)$ is applied to the different patches, there will be correlations between outputs. For this reason, we consider the mapping $\vf(\cdot)$ a multi-output GP (MOGP), and name it the Multi-Output Convolutional Kernel (MOCK). 
Multi-output GPs \citep{alvarez2012kernels} can be characterised by their covariance between the different outputs $f_{p}$ and $f_{q}$ of different inputs $\vx$ and $\vx'$, giving in our case
\begin{align}
    \cov\left[f_{p}(\vx), f_{q}(\vx')\right] = k_g\left(\vx^{[p]}, \vx'^{[q]}\right).
\end{align}
In this setting, if we are dealing with $N$ images of $P$ patches, the corresponding covariance matrix has a size of $N\!\times\!N\!\times\!P\!\times\!P$, which makes its calculation and inversion infeasible for most datasets.  %

Efficient inference for MOGPs relies strongly on choosing useful inducing variables. We developed a framework for generic MOGPs that allows for the flexible specification of both multi-output priors and inducing variables. This means that we can take computational advantage of independence properties of the prior. Given our framework, which puts the right mathematical and software abstractions in place, the implementation of a complex MOGP, such as a DCGP, is not much more difficult than that of a single-output GP \citep{gpflow2}.

\section{VARIATIONAL INFERENCE WITH SPARSE GAUSSIAN PROCESSES}
\label{sec:inference}

Consider a training dataset $\{(\vx_n, y_n)\}_{n=1}^N \subset \Re^D \times \Re$, consisting of $N$ images $\vx_n \in \Re^D$ and class labels $y_n$. We set up a deep convolutional GP as $f(\cdot) := f_L(\ldots f_2(f_1(\cdot)))$, where each
\begin{equation*}
    f_\ell(\cdot) \sim \GP\big(0, k_\ell(\cdot, \cdot)\big)\text{ and }
    y_n \given f, \vx_n \sim p\big(y_n \given f(\vx_n)\big).
\end{equation*}
We refer to the latent function-evaluation of a hidden GP as $\vh_{n, \ell} = f_\ell(\vh_{n, \ell-1})$ and, for convenience, we define $\vh_{n, 0} := \vx_n$. We assume that each function is a MOGP with $P_\ell$ (correlated) outputs.

Given this setup, we are interested in both the posterior $p(f(\cdot) \given \vy)$ for making subsequent predictions and the marginal likelihood (evidence) $p(\vy)$ to optimise the model's hyperparameters. Exact inference is not possible in this setting given our non-conjugate likelihood $p(y_n \given \vh_{n, L})$ and the $\BigO\left(N^3\right)$ cost of operations on covariance matrices, limiting the size of the datasets.

We use sparse variational GPs to address these issues, following \citet{titsias2009}, \citet{hensman2013}, and \citet{matthews16}. The framework conditions the prior on \emph{inducing variables} $\vu_\ell$, and then specifies a free Gaussian density $q(\vu_\ell) = \Gauss(\vm_\ell, \MS_\ell)$. This gives the approximation $q(f_\ell(\cdot)) = \int p(f_\ell(\cdot) \given \vu_\ell)\, q(\vu_\ell) \diff \vu_\ell$ for each layer. The original framework chose the inducing outputs $\vu_\ell$ to be observations of the GP to some inducing inputs $\MZ_\ell = \{\vz_{\ell, m}\}_{m=1}^M$, i.e. $\vu_\ell = f_\ell(\MZ_\ell)$. 
Even though we are representing the GP at a finite set of points, the posterior is still a full-rank GP. It predicts using an infinite number of basis functions thanks to the use of the prior conditional.
The overall approximate posterior has the form $q(f_\ell(\cdot)) = \GP (\mu_{\ell}(\cdot), \nu_{\ell}(\cdot))$ with
\begin{align}
\label{eq:qf}
   \mu_{\ell}(\cdot) &= \vk_{\vu_\ell}^\top(\cdot) \MK_{\vu_\ell\vu_\ell}^{-1} \vm_\ell\\
   \nu_{\ell}(\cdot) &= k_\ell(\cdot, \cdot) + \vk_{\vu_\ell}^\top(\cdot) \MK_{\vu\vu_\ell}^{-1}(\MS_\ell - \MK_{\vu\vu_\ell})\MK_{\vu\vu_\ell}^{-1} \vk_{\vu_\ell}(\cdot),\nonumber
\end{align}
where $\vm_\ell \in \Re^M$, and $\MS_\ell \in \Re^{M\times M}$ are variational parameters to be learned by optimisation. When we predict for a single point, the size of $\vk_{\vu_\ell}(\cdot)$ is $P_\ell \times M$ (that is, the number of outputs by the number of inducing variables), while $k_\ell(\cdot, \cdot)$ returns the $P_\ell\times P_\ell$ covariance matrix for all outputs. Crucially, because we are dealing with MOGPs, our posterior mean $\mu_\ell(\cdot)$ has size $\Re^{P_\ell}$ and $\nu_{\ell}(\cdot) \in \Re^{P_\ell \times P_\ell}$ grows quadratically in the number of outputs, roughly corresponding to the number of input pixels.

Following the standard variational \citep{hensman2013, hoffman2013} approach, we construct a lower bound to the marginal likelihood (known as the Evidence Lower BOund, or ELBO) which we then optimise to find the optimal approximate posterior and the model's hyperparameters.
To derive the ELBO, we start with the joint density for the generative model
\begin{multline*}
    p(\{y_n\}_n, \{\vh_{n, \ell}\}_{n, \ell}, \{f_\ell(\cdot)\}_\ell) =\\
    \prod_n p(y_n \given \vh_{n, L}) \prod_{\ell} p(\vh_{n, \ell} \given \vh_{n, \ell-1}, f_\ell(\cdot))~p(f_\ell(\cdot)),
\end{multline*}
and an approximate variational posterior $q(\{\vh_{n, \ell}\}_{n, \ell}, \{f_\ell(\cdot)\}_\ell)$ which we give the form $\prod_{n=1}^N \prod_{\ell=1}^L p(\vh_{n, \ell} \given \vh_{n, \ell-1}, f_\ell(\cdot))\,q(f_\ell(\cdot))$.
The repetition of $p(\vh_{n, \ell} \given \vh_{n, \ell-1}, f_\ell(\cdot))$ in both the prior and posterior leads to their cancellation in the lower bound
\begin{multline}
\log p(\vy) \ge
  \sum\nolimits_n \E_{q(\vh_{n, L})}
    \left[\log p(y_n \given \vh_{n, L}) \right]\\
 - \sum\nolimits_{\ell} \KL \left[q(\vu_\ell) || p(\vu_\ell) \right].  \label{eq:elbo}
\end{multline}
The form of $p(\vh_{n, \ell} \given \vh_{n, \ell-1}, f_\ell(\cdot))$ leads to different DGP models. \citet{damianou2013deep} used a Gaussian distribution $\Gauss(\vh_{n, \ell} \given f_\ell(\vh_{n, \ell-1}), \sigma^2_\ell)$, which requires an additional approximate posterior over the $\vh_\ell$'s in the bound. We follow \citet{salimbeni2017doubly} and use a deterministic map between $\vh_{n, \ell}$ and $\vh_{n, \ell-1}$ given $f_\ell(\cdot)$, corresponding to $\delta\left\{\vh_{n, \ell} = f_\ell(\vh_{n, \ell-1})\right\}$.

We can obtain an unbiased estimate of \cref{eq:elbo} by only considering a random subset of the training data to cheaply estimate the first term and by rescaling the KL term appropriately. The expectation over $q(\{\vh_{n, \ell}\}_{n, \ell}, \{f_\ell(\cdot)\}_\ell)$ is evaluated using Monte-Carlo, %
see \citet{salimbeni2017doubly} for details.

\subsection{Computational Complexity of Dealing with Correlated Convolutional Layers}
\label{sec:computation}
To evaluate the expectation in the ELBO as described above, we need to generate samples of $q(f_\ell(\cdot))$ with the covariance $\nu_\ell(\vh_{n,\ell-1})$. This requires taking a Cholesky of this covariance, of which we have one for each datapoint in the minibatch. This presents a significant computational problem, because its size is $P_\ell\times P_\ell$, with $P_\ell$ being roughly the same as the number of patches in the input image. Compared to a non-convolutional deep GP, where we only need a single Cholesky for each layer of $\MK_{\vu_\ell\vu_\ell}$, this adds a large computational cost.
The deep convolutional GP model of \citet{blomqvist2019deep} suffers from this problem as well. Their method avoids this computational cost by simply sampling from the $P_\ell$ marginals, ignoring the between-patch correlation. 
In the supplementary material (see \cref{fig:sampling-full-cov}) we study the difference between both approaches and find that the lower computation cost, $\mathcal{O}(P_\ell)$, of sampling from the marginals drastically improves the number of iterations per second and is worth the minor reduction in performance. The bias introduced to the gradient of the ELBO appears to have little effect.

\subsection{Inter-Domain Inducing Patches}

So far, we have set up the optimisation objective (\cref{eq:elbo}) and defined the approximate posterior GP for each layer $q(f_\ell(\cdot))$. %
The final issues we need to address are (1) the impractically large double sums over all patches for computing entries of the $\MK_{\vu_\ell\vu_\ell}$ and (2) the organisational complexity of dealing with inducing variables multi-output $\vu_\ell$ in MOGP.

Using inter-domain inducing variables \citep{lazaro2009inter} solves the mathematical, organisational, and software problems of both issues. We follow \Citet{van2017convolutional} to define for each layer $\vu_\ell$ as evaluations of the patch response function $g_{\ell}(\cdot)$, and we place the inducing inputs in the patch space $\Re^{wh}$, rather than image space $\Re^{P_{\ell-1}}$. The GP inter-domain and multi-output software framework, which we codeveloped with this work, enables us to implement this in an efficient and modular way \citep{gpflow2}.

To apply this approximation  in \eqref{eq:qf} and implement this in the framework, we need to find $\vk_{\vu_\ell}(\cdot)$ and $\MK_{\vu_\ell \vu_\ell}$:
\begin{align*}
    \vk_{\vu_\ell}(\vh_{n, \ell-1}) &= \E [ g_\ell (\MZ_\ell) f_\ell(\vh_{n, \ell-1})] = \left[k_{g_\ell}(\MZ_\ell, \vh_{n, \ell-1}^{[p]})\right]_{p=1}^{P_\ell}\\
    \MK_{\vu_\ell \vu_\ell} &= %
    k_{g}(\MZ_\ell, \MZ_\ell).
\end{align*}
Choosing the inducing variables in this way greatly reduces the computational cost of the method, because we now require covariances only between the patches of the input image and the inducing patches. More precisely, $\MK_{\vu_\ell \vu_\ell}$ and $\vk_{\vu_\ell}(\vh_{n, \ell-1})$ become $M \times M$ and $M \times P_\ell$ sized tensors.

\section{EXPERIMENTS}

\begin{table*}[t]
    \centering
    \caption{Results of classification experiments with Bayesian CNN (B-CNN) and shallow GP models. We report the top-$k$ error rate, and Negative Log-Likelihood (NLL) on the full test set and on the misclassified images of the test set. The TICK-GP outperforms the Conv-GP on every dataset in both accuracy and NLL, illustrating the clear benefits of translation insensitivity. The single-layer TICK-GP models get a similar accuracy to the CNNs but have more calibrated predictive probabilities (lower NLL). \label{tab:classification}}
    
    \vspace{.3cm}
    \scalebox{.8}{\begin{tabular}{l llll l llll l llll}
\toprule
        & \multicolumn{4}{c}{MNIST} &  & \multicolumn{4}{c}{FASHION-MNIST} & & \multicolumn{4}{c}{GREY CIFAR-10} \\ \cmidrule{2-5} \cmidrule{7-10} \cmidrule{12-15}
metric             & SE   & Conv    & TICK          & B-CNN            & & SE   & Conv  & TICK         & B-CNN          &  & SE       & Conv  & TICK         & B-CNN              \\
\midrule                                                                                                                                                         
top-$1$ error (\%)     & 2.31     & 1.70        &  \textbf{0.83}  & 0.91       & & 12.15     & 11.06       & 10.01       & \textbf{8.31}   &  & 58.24        & 41.65   & 37.82       & \textbf{37.44}    \\
top-$2$ error (\%)     & 0.69     & 0.49        & \textbf{0.11}   & 0.22       & & 3.67      & 3.18        & 2.69        & \textbf{2.17}   &  & 38.91        & 24.09   & \textbf{20.52}  & {21.48}    \\
top-$3$ error (\%)     & 0.35     & 0.19        & \textbf{0.05}   & \textbf{0.05}  & & 1.21      & 1.11        & 0.92        & \textbf{0.75}   &  & 27.18        & 14.93   & \textbf{12.21}  & {12.91}    \\
NLL full ($\times 10$)  & 0.60     & 0.57        & \textbf{0.29}   & 0.35       & & 2.68      & 2.52        & \textbf{2.28}   & 2.45       &  & 15.56        & 11.68   & \textbf{10.56}  & 11.16        \\
NLL misses             & 1.86     & 1.97        & \textbf{1.70}   & 2.58       & & 1.90      & 1.90        & \textbf{1.89}   & 2.10       &  & 2.20         & 2.12    & \textbf{2.10}   & 2.23        \\
\bottomrule
\end{tabular}

}
\end{table*}

\begin{figure}[th]
	\includegraphics[width=0.9\linewidth]{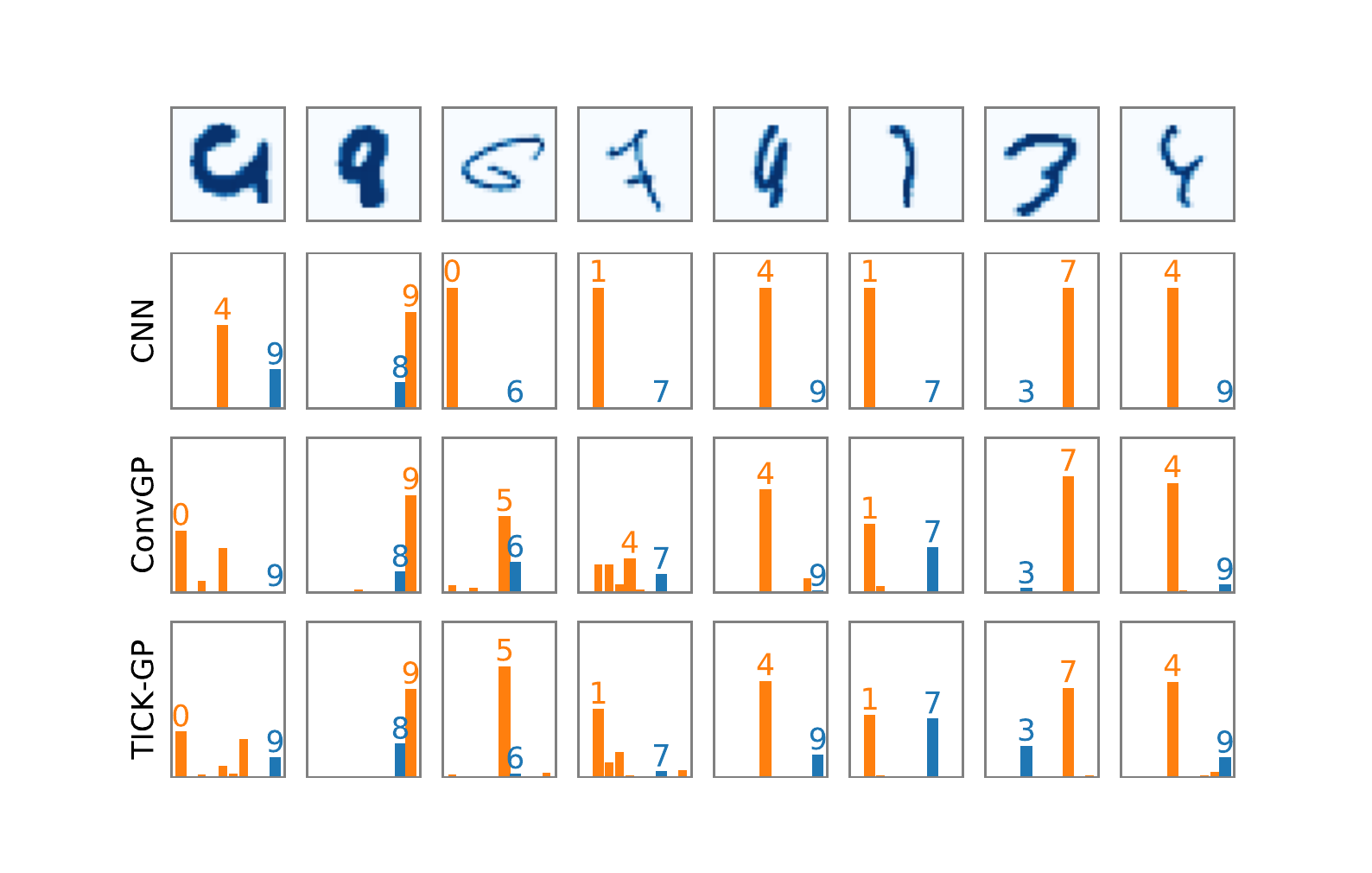}
	\caption{Posterior prediction probabilities for a (random) subset of \emph{misclassified} images (top row) from MNIST. The barplot shows the probabilities for each of the classes, 0 to 9. The largest (orange) bar is the model's prediction. The blue bar is the true class label. The CNN predicts the wrong classes with high certainty, while the GP quantifies uncertainty better.}
	\label{fig:mnist-posterior-probs}
\end{figure}

We present results using our TICK in deep and shallow GP models. First, we show that TICK-GP improves over Conv-GP and achieves the highest reported classification result for GPs on standard classification tasks in terms of accuracy and calibration. Secondly, we compare our method with Bayesian CNNs and find that TICK-GP's uncertainty estimates are superior, and that the ELBO can be used for model selection and automated training. We show that the CNN is confidently wrong on some ambiguous cases, while TICK-GP provides calibrated uncertainty. In \cref{sec:appendix:ood}, we demonstrate that this effect is even more pronounced in a transfer learning task. In \cref{sec:exp:deep}, we also show the benefits of translation insensitivity in deep GPs.

\subsection{Comparison to Conv-GP and Bayesian Neural Nets on Image Classification}
\label{sec:exp:mnist}

We evaluate TICK-GP on three image benchmarks (MNIST, FASHION-MNIST, and grey-scale CIFAR-10) and compare its performance to a SE-GP, Conv-GP \citep{van2017convolutional} and a Bayesian CNN (B-CNN) based on dropout \citep{gal2016dropout}.  All GP models in this experiment are single-layered and trained following the method outlined in \cref{sec:inference}. Their exact setup (kernel, number of inducing points, learning rate schedule, etc.) is detailed in \cref{sec:appendix:setup-model}. We compare the GP models to a Bayesian CNN architecture, with two convolutional layers followed by two full dense layers. We use dropout at train and test time, following \citep{gal2016dropout}, with 50\% keep-probability, which we found by running a grid search and selecting the best model based on its NLL on a 10\% validation set.
We further detail the CNN configuration in \cref{sec:appendix:cnn}. 
The SE-GP is a vanilla Sparse Variational GP (SVGP) \citep{hensman2013} using a SE kernel. %

\Cref{tab:classification} reports the top-$k$ error rate and the Negative Log-Likelihood (NLL). We use NLL as our main metric for calibration because it is a proper scoring rule \citep{gneiting2007strictly} and has a useful relationship to returns obtained from bets on the future based on the predicted belief \citep{roulston2002evaluating}. The top-$k$ error rate is the percentage of test images for whom the true class label is not within the highest $k$ predictive probabilities. We see that TICK-GP outperforms previous GP models and dropout-based CNNs in terms of NLL, both on the complete test set and on the misclassified images. The single-layer TICK-GP sets the new records of classification with GP models on the listed datasets, showing the importance of encoding the right inductive biases into a GP model. Most importantly, the model is comparable with the B-CNN in terms of error rate, but has better-calibrated predictive probabilities, which enables ELBO-based model selection, as shown in the next section.

\Cref{fig:mnist-posterior-probs} shows the predictive probability for a few randomly selected misclassified images, demonstrating both the better-calibrated probabilities of GP-based models compared to the CNN models, and the improvements of the new TICK-GP over the Conv-GP. We clearly see how the CNN can be very confidently wrong. In \cref{sec:appendix:classification-images} we show the complete set of misclassified images.

\begin{figure}[bt]
	\includegraphics[width=1\linewidth]{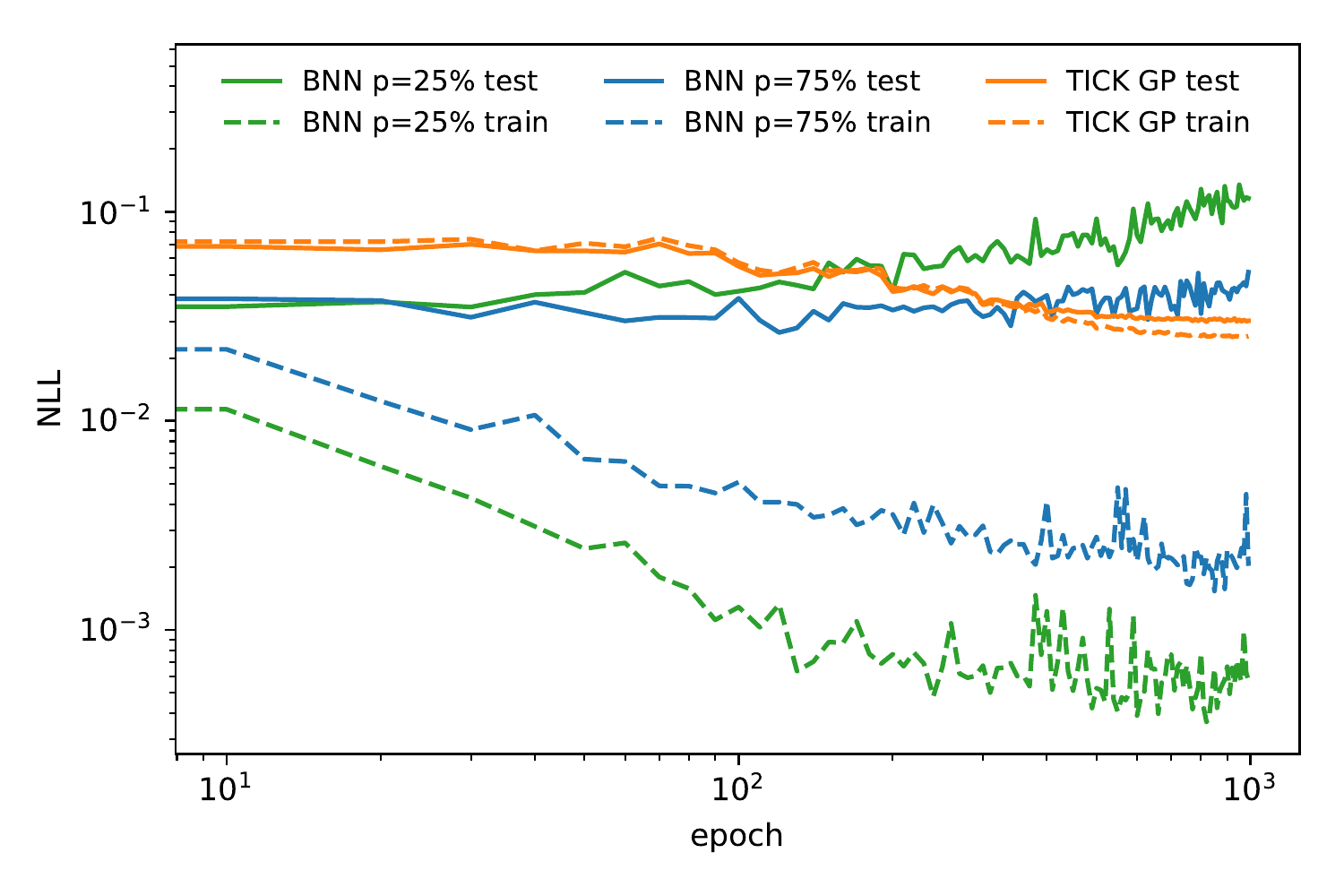}
	\caption{B-CNN marginal likelihood estimates are not usable for hyperparameter selection. The plot shows the train and test NLL of two B-CNNs w.r.t. epochs on MNIST. We notice that both models overfit, shown by the gap between train and test performance, and the deteriorating test performance. A higher dropout rate ($p=75\%$ vs. $25\%$) postpones this effect, but does not prevent it. For the TICK-GP model we see that the train NLL is a good proxy for the test NLL, and that it outperforms the B-CNNs over time. \label{fig:dropout}}
\end{figure}

\subsection{Comparison to Bayesian Neural Networks with different Dropout Rates}
\label{sec:exp:dropout}

While Bayesian deep learning methods based on dropout \citep{gal2016dropout} have been empirically successful in improving the quality of uncertainty estimates, it is unclear to what extent they accurately approximate the true posteriors \citep{hron2018dropout}. Additionally, in this experiment we find that they do not provide a Bayesian objective that allows for the automated training of hyperparameters and model selection \citep{rasmussen2001occam}. In this experiment we have not considered other Bayesian deep learning approaches like \citep{lee2018, khan2019}.

\Cref{fig:dropout} shows how powerful the marginal likelihood (ELBO) of our fully Bayesian model is. In the plot, we show the training traces of a TICK-GP and two B-CNNs with different dropout rates on MNIST. The models have the same setup as in \cref{sec:exp:mnist}. We notice the gap between the train and test NLL of the B-CNNs, and a train NLL which keeps decreasing while the NLL of the test set starts to increase.
Using larger dropout rates ($p_\text{dropout}=0.75$ instead of $0.25$) postpones this effect but does not prevent it.
By contrast, the proper marginal likelihood objective of the TICK-GP model is reflected by the close similarity of the test NLL and the train NLL. This enables us to do automated model selection through higher marginal likelihood and ELBO-based hyperparameter learning (e.g.~ to learn the degree of translation insensitivity of a convolutional layer).

\subsection{Translation Insensitivity in Deep Convolutional GPs (DCGPs)}
\label{sec:exp:deep}

Having shown how TICK-GP compares to vanilla Conv-GPs, we now consider deep architectures. In \cref{tab:deep} we list the performance of a deep Conv-GP (DCGP)  \citep{blomqvist2019deep} and a deep TICK-GP (ours) on MNIST and CIFAR-10.  
We configure the models as identically as possible: each layer uses 384 inducing 5x5 patches (initialised using random patches from the training images), an identity Conv2D mean function for the hidden layers, and a SE kernel for the patch response function (the complete setup can be found in \cref{sec:appendix:setup-model}).

The deep TICK-GP, which can learn the degree of insensitivity, outperforms the plain DCGP in terms of accuracy and NLL for any depth. We see that both models improve with depth, and more importantly, that the ELBO is reflecting this. This can also be observed in the appendix (\cref{fig:appendix:traces-dgp}), where deep TICK-GPs are consistently outperforming deep convolutional GPs. We also compare to a growing dropout-based B-CNN, which gives for the top-1 error and NLL: 1.93\%, 0.07 (1 layer), 1.04\%, 0.03 (2 layers) 0.86\%, 0.04 (3 layers) on MNIST. As expected, the B-CNN’s accuracy improves with depth, but the NLL (uncertainty quantification) gets worse. This is in contrast with our model which continues to improve NLL and accuracy with depth. 

To further position the TICK-GP, we compare its performance against non-convolutional deep GPs. On MNIST we found that a deep GP \citep{salimbeni2017doubly} with SE kernels, 2 layers, and 384 inducing inputs per layer managed 98\% accuracy. This is equal to a vanilla GP classifier (\citet{hensman2015scalable} report 98\% accuracy), illustrating that depth by itself does not always improve performance when the wrong inductive biases are encoded in the model. \citet{havasi2018inference} report similar conclusions in their work; their HMC approach delivers 98.0\% accuracy.  Our model beats all of these methods with 99.33\% accuracy.
Both non-convolutional deep GP papers do not report results for CIFAR. We ran the method of \citet{salimbeni2017doubly} and managed to get 47.26\% accuracy. Our model outperforms this with 74.41\% accuracy.

\begin{table}[t]
\centering
\caption{DCGP \citep{blomqvist2019deep} (reproduced with our code) and Deep TICK-GP (our method) on MNIST and CIFAR-10. \label{tab:deep}}
\scalebox{0.8}{\begin{tabular}{ll ll l ll}
\toprule
            &        & \multicolumn{2}{c}{MNIST} &  & \multicolumn{2}{c}{CIFAR-10} \\ \cmidrule{3-4} \cmidrule{6-7}
depth    & metric & Conv & TICK & &  Conv & TICK  \\
\midrule
1   & top-1 error (\%)               	& 1.87  & \textbf{1.19}  &  & 41.06 & \textbf{37.10} \\
        & NLL full            			& 0.06 & \textbf{0.04} &  & 1.17 & \textbf{1.08} \\
        & neg. ELBO ($\times10^3$)      & 8.29 & \textbf{5.83}   &  & 65.72 & \textbf{63.51} \\
        \cmidrule{2-7}
2   & top-1 error (\%)                & 0.96  & \textbf{0.67}  &  & 28.60 & \textbf{25.59} \\
        & NLL full        				& 0.04 & \textbf{0.02} &  & 0.84 & \textbf{0.75} \\
        & neg. ELBO ($\times10^3$)  		& 5.37  & \textbf{4.25}  &  & 52.81 & \textbf{48.31} \\
        \cmidrule{2-7}
3   & top-1 error (\%)               	& 0.93       & \textbf{0.64}  &  & 25.33 & \textbf{23.83} \\
        & NLL full          			& 0.03      & \textbf{0.02} &  & 0.74 & \textbf{0.69} \\
        & neg. ELBO ($\times10^3$)  		& 5.045       & \textbf{4.19}  &  & 49.38 & \textbf{47.53} \\
\bottomrule
\end{tabular}
}
\end{table} 

\subsection{Implementation and Reproducibility}

The main implementation difficulty for multi-output GPs is dealing with a large amount of special cases to ensure the most efficient code path is used. This makes it a challenge to implement modular and reusable code; the correct software abstractions should be used to keep the code readable, manageable, and extendable.
We noticed, however, that most multi-output GPs \citep{alvarez2012kernels} can be reformulated in terms of single-output inducing outputs, leaning towards inter-domain approximations. Based on this observation we developed a general multi-output GP framework \citep{gpflow2}.

To implement our model in the framework we need to provide components specific to our model. In particular, we need to implement the multi-output and single-output convolutional kernels (\cref{eq:conv-gp} and \cref{eq:deconv-f-def}) and the corresponding single and multi-output approximate posterior GP $q(f(\cdot))$ (different flavours of \cref{eq:qf}). Finally, we also need to implement the bound in \cref{eq:elbo}.

\section{CONCLUDING DISCUSSION}
We have shown that the accuracy of Bayesian methods, and the quality of their posterior uncertainties, depends strongly on the suitability of the modelling assumptions made in the prior, and that Bayesian inference by itself is often not enough. This motivated us to develop the Translation Insensitive Convolutional Kernel  (TICK), which leads to improved uncertainty estimates and accuracy on a range of different problems, and sets the new state-of-art results for GP models.

While we appreciate that our experiments are still on rudimentary image datasets (e.g.~not of the calibre of ImageNet), they do show that when the accuracy of our method is on a par with that of a neural network, we outperform the neural network in terms of uncertainty estimation (\cref{sec:exp:mnist}) and in the use of marginal likelihood approximations for hyperparameter learning (\cref{sec:exp:dropout}). We believe that this suggests that the full benefits of the Bayesian framework are not currently realised by Bayesian deep learning. 

We further presented deep convolutional GPs in a new and clear way: image-to-image layers modelled as correlated multi-output GPs (\cref{sec:dcgp}). This enabled efficient implementation in our general-purpose open-sourced framework \citep{gpflow2}. We also highlighted a computational limitation of current convolutional layers (\cref{sec:computation}) which had not been addressed in earlier work, and which future work should focus on.

\subsection*{Acknowledgements}
We have greatly appreciated valuable discussions with Marc Deisenroth and Zhe Dong in the preparation of this work. We would like to thank Fergus Simpson, Hugh Salimbeni, ST John, Victor Picheny, and anonymous reviewers for helpful feedback on the manuscript.

\bibliographystyle{plainnat}
\bibliography{bib.bib}

\begin{thebibliography}{33}
\providecommand{\natexlab}[1]{#1}
\providecommand{\url}[1]{\texttt{#1}}
\expandafter\ifx\csname urlstyle\endcsname\relax
  \providecommand{\doi}[1]{doi: #1}\else
  \providecommand{\doi}{doi: \begingroup \urlstyle{rm}\Url}\fi

\bibitem[Adler et~al.(1981)Adler, Firmin, and Kendall]{adler198}
Robert~J. Adler, D.~Firmin, and David~George Kendall.
\newblock A non-{G}aussian model for random surfaces.
\newblock \emph{Philosophical Transactions of the Royal Society of London.
  Series A, Mathematical and Physical Sciences}, 303\penalty0 (1479):\penalty0
  433--462, 1981.

\bibitem[Alvarez et~al.(2012)Alvarez, Rosasco, and
  Lawrence]{alvarez2012kernels}
Mauricio~A. Alvarez, Lorenzo Rosasco, and Neil~D. Lawrence.
\newblock Kernels for vector-valued functions: A review.
\newblock \emph{Foundations and Trends in Machine Learning}, 2012.

\bibitem[Blomqvist et~al.(2019)Blomqvist, Kaski, and
  Heinonen]{blomqvist2019deep}
Kenneth Blomqvist, Samuel Kaski, and Markus Heinonen.
\newblock Deep convolutional {G}aussian {P}rocesses.
\newblock In \emph{European Conference on Machine Learning and Principles and
  Practice of Knowledge Discovery in Databases}, 2019.

\bibitem[Damianou and Lawrence(2013)]{damianou2013deep}
Andreas Damianou and Neil~D. Lawrence.
\newblock {Deep {G}aussian {P}rocesses}.
\newblock In \emph{Artificial Intelligence and Statistics}, 2013.

\bibitem[Gal and Ghahramani(2016)]{gal2016dropout}
Yarin Gal and Zoubin Ghahramani.
\newblock Dropout as a {B}ayesian approximation: Representing model uncertainty
  in deep learning.
\newblock In \emph{International Conference on Machine Learning}, 2016.

\bibitem[Ghafoorian et~al.(2017)Ghafoorian, Karssemeijer, Heskes, van Uden,
  Sanchez, Litjens, de~Leeuw, van Ginneken, Marchiori, and
  Platel]{ghafoorian2017location}
Mohsen Ghafoorian, Nico Karssemeijer, Tom Heskes, Inge W.~M. van Uden, Clara~I
  Sanchez, Geert Litjens, Frank-Erik de~Leeuw, Bram van Ginneken, Elena
  Marchiori, and Bram Platel.
\newblock Location sensitive deep convolutional neural networks for
  segmentation of white matter hyperintensities.
\newblock \emph{Scientific Reports}, 7\penalty0 (1):\penalty0 5110, 2017.

\bibitem[Gneiting and Raftery(2007)]{gneiting2007strictly}
Tilmann Gneiting and Adrian~E. Raftery.
\newblock Strictly proper scoring rules, prediction, and estimation.
\newblock \emph{Journal of the American Statistical Association}, 102\penalty0
  (477):\penalty0 359--378, 2007.

\bibitem[Goodfellow et~al.(2016)Goodfellow, Bengio, Courville, and
  Bengio]{goodfellow2016deep}
Ian Goodfellow, Yoshua Bengio, Aaron Courville, and Yoshua Bengio.
\newblock \emph{Deep learning}, volume~1.
\newblock MIT press Cambridge, 2016.

\bibitem[Havasi et~al.(2018)Havasi, Hern{\'a}ndez-Lobato, and
  Murillo-Fuentes]{havasi2018inference}
Marton Havasi, Jos{\'e}~Miguel Hern{\'a}ndez-Lobato, and Juan~Jos{\'e}
  Murillo-Fuentes.
\newblock {I}nference {I}n deep {G}aussian {P}rocesses using stochastic
  gradient {H}amiltonian {M}onte {C}arlo.
\newblock In \emph{Advances in Neural Information {P}rocessing Systems}, pages
  7506--7516, 2018.

\bibitem[Hensman et~al.(2013)Hensman, Fusi, and Lawrence]{hensman2013}
James Hensman, Nicolo Fusi, and Neil~D. Lawrence.
\newblock {{G}aussian {P}rocesses for Big Data}.
\newblock \emph{Uncertainty in Artificial Intelligence}, 2013.

\bibitem[Hensman et~al.(2015)Hensman, Matthews, and
  Ghahramani]{hensman2015scalable}
James Hensman, Alexander G. de~G. Matthews, and Zoubin Ghahramani.
\newblock Scalable variational {G}aussian {P}rocess {C}lassification.
\newblock In \emph{Artificial Intelligence and Statistics}, 2015.

\bibitem[Hoffman et~al.(2013)Hoffman, Blei, Wang, and Paisley]{hoffman2013}
Matthew~D. Hoffman, David~M. Blei, Chong Wang, and John Paisley.
\newblock {Stochastic Variational Inference}.
\newblock \emph{Journal of Machine Learning Research}, 2013.

\bibitem[Hron et~al.(2018)Hron, Matthews, and Ghahramani]{hron2018dropout}
Jiri Hron, Alexander G. de~G. Matthews, and Zoubin Ghahramani.
\newblock Variational {B}ayesian dropout: pitfalls and fixes.
\newblock In \emph{International Conference on Machine Learning}, 2018.

\bibitem[Huang et~al.(2017)Huang, Liu, Van Der~Maaten, and
  Weinberger]{densenet}
Gao Huang, Zhuang Liu, Laurens Van Der~Maaten, and Kilian~Q. Weinberger.
\newblock Densely connected convolutional networks.
\newblock In \emph{IEEE conference on computer vision and pattern recognition},
  2017.

\bibitem[Keras()]{cnn2017tf}
Keras.
\newblock Keras implementation of {CNN} for {MNIST}.
\newblock Available from
  \url{https://github.com/keras-team/keras/blob/master/examples/mnist_cnn.py}.

\bibitem[Kingma and Ba(2014)]{kingma2014adam}
Diederik~P. Kingma and Jimmy Ba.
\newblock {Adam}: A method for stochastic optimization.
\newblock In \emph{International Conference on Learning Representations}, 2014.

\bibitem[L{\'a}zaro-Gredilla and Figueiras-Vidal(2009)]{lazaro2009inter}
Miguel L{\'a}zaro-Gredilla and An{\'\i}bal Figueiras-Vidal.
\newblock Inter-domain {{G}aussian} {{P}rocesses} for sparse inference using
  inducing features.
\newblock In \emph{Neural Information {P}rocessing Systems}, 2009.

\bibitem[LeCun et~al.(1989)LeCun, Boser, Denker, Henderson, Howard, Hubbard,
  and Jackel]{lecun1989backpropagation}
Yann LeCun, Bernhard Boser, John~S. Denker, Donnie Henderson, Richard~E.
  Howard, Wayne Hubbard, and Lawrence~D. Jackel.
\newblock Backpropagation applied to handwritten zip code recognition.
\newblock \emph{Neural Computation}, 1\penalty0 (4):\penalty0 541--551, 1989.

\bibitem[Lee et~al.(2018)Lee, Sohl-dickstein, Pennington, Novak, Schoenholz,
  and Bahri]{lee2018}
Jaehoon Lee, Jascha Sohl-dickstein, Jeffrey Pennington, Roman Novak, Sam
  Schoenholz, and Yasaman Bahri.
\newblock Deep neural networks as {G}aussian {P}rocesses.
\newblock In \emph{International Conference on Learning Representations}, 2018.

\bibitem[Liu et~al.(2018)Liu, Lehman, Molino, Such, Frank, Sergeev, and
  Yosinski]{liu2018}
Rosanne Liu, Joel Lehman, Piero Molino, Felipe~Petroski Such, Eric Frank, Alex
  Sergeev, and Jason Yosinski.
\newblock An intriguing failing of convolutional neural networks and the
  coordconv solution.
\newblock In \emph{Neural Information {P}rocessing Systems}, 2018.

\bibitem[Matthews et~al.(2016)Matthews, Hensman, Richard, and
  Ghahramani]{matthews16}
Alexander G. de~G. Matthews, James Hensman, Turner Richard, and Zoubin
  Ghahramani.
\newblock {On Sparse Variational Methods and the Kullback-Leibler Divergence
  between Stochastic {P}rocesses}.
\newblock \emph{Artificial Intelligence and Statistics}, 2016.

\bibitem[Neal(1996)]{neal1996bayesian}
Radford~M. Neal.
\newblock \emph{Bayesian learning for neural networks}, volume 118.
\newblock Springer, 1996.

\bibitem[Osawa et~al.(2019)Osawa, Swaroop, Khan, Jain, Eschenhagen, Turner, and
  Yokota]{khan2019}
Kazuki Osawa, Siddharth Swaroop, Mohammad Emtiyaz~E Khan, Anirudh Jain, Runa
  Eschenhagen, Richard~E Turner, and Rio Yokota.
\newblock Practical deep learning with bayesian principles.
\newblock In \emph{Advances in Neural Information {P}rocessing Systems 32},
  pages 4287--4299. Curran Associates, Inc., 2019.

\bibitem[Rasmussen and Ghahramani(2001)]{rasmussen2001occam}
Carl~E. Rasmussen and Zoubin Ghahramani.
\newblock Occam's {R}azor.
\newblock In \emph{Neural Information {P}rocessing Systems}. 2001.

\bibitem[Rasmussen and Williams(2006)]{rasmussen2006}
Carl~E. Rasmussen and Christopher K.~I. Williams.
\newblock \emph{{{G}aussian {P}rocesses for Machine Learning}}.
\newblock MIT Press, 2006.

\bibitem[Roulston and Smith(2002)]{roulston2002evaluating}
Mark~S. Roulston and Leonard~A. Smith.
\newblock Evaluating probabilistic forecasts using information theory.
\newblock \emph{Monthly Weather Review}, 130\penalty0 (6):\penalty0 1653--1660,
  2002.

\bibitem[Salimbeni and Deisenroth(2017)]{salimbeni2017doubly}
Hugh Salimbeni and Marc~P. Deisenroth.
\newblock {Doubly Stochastic Variational Inference for Deep {G}aussian
  {P}rocesses}.
\newblock \emph{Neural Information {P}rocessing Systems}, 2017.

\bibitem[Szegedy et~al.(2017)Szegedy, Ioffe, Vanhoucke, and Alemi]{inception}
Christian Szegedy, Sergey Ioffe, Vincent Vanhoucke, and Alexander~A. Alemi.
\newblock Inception-v4, inception-resnet and the impact of residual connections
  on learning.
\newblock In \emph{AAAI Conference on Artificial Intelligence}, 2017.

\bibitem[Titsias(2009)]{titsias2009}
Michalis Titsias.
\newblock {Variational Learning of Inducing Variables in Sparse {G}aussian
  {P}rocesses}.
\newblock \emph{Artificial Intelligence and Statistics}, 2009.

\bibitem[UCI()]{semeion2008}
UCI.
\newblock Semeion handwritten digit data set.
\newblock Available from
  \url{https://archive.ics.uci.edu/ml/datasets/semeion+handwritten+digit}.

\bibitem[van~der Wilk et~al.(2017)van~der Wilk, Rasmussen, and
  Hensman]{van2017convolutional}
Mark van~der Wilk, Carl~E. Rasmussen, and James Hensman.
\newblock Convolutional {{G}aussian} {{P}rocesses}.
\newblock In \emph{Neural Information {P}rocessing Systems}, 2017.

\bibitem[van~der Wilk et~al.(2018)van~der Wilk, Bauer, John, and
  Hensman]{vdw2018invariances}
Mark van~der Wilk, Matthias Bauer, ST~John, and James Hensman.
\newblock {Learning Invariances using the Marginal Likelihood}.
\newblock In \emph{Neural Information {P}rocessing Systems}, 2018.

\bibitem[{van der Wilk} et~al.(2020){van der Wilk}, Dutordoir, John, Artemev,
  Adam, and Hensman]{gpflow2}
Mark {van der Wilk}, Vincent Dutordoir, ST~John, Artem Artemev, Vincent Adam,
  and James Hensman.
\newblock {A} {F}ramework for {I}nterdomain and {M}ultioutput {G}aussian
  {P}rocesses.
\newblock \emph{arXiv preprint arXiv:2003.01115}, 2020.

\end{thebibliography}

\begin{appendices}
  \crefalias{section}{appsec}

\onecolumn
\clearpage

\section{EXPERIMENT SETUP}
\label{sec:appendix:setup-model}

\subsection{MNIST, FASHION-MNIST and CIFAR-10 Experiment (\cref{sec:exp:mnist})}
The SE-GP model is a vanilla Sparse Variational GP (SVGP) \citep{hensman2013} using a SE kernel defined directly on the images. For all datasets, we use the standard splits for the train and test set, as returned by the Keras dataset library. For comparison's sake, we set up TICK-GP and Conv-GP in as similar way as possible. They are both configured to have 1,000 inducing 5x5 patches, which are initialised using randomly picked patches from the training examples.
We choose a SE kernel for the patch response function, and follow \citet{van2017convolutional} in multiplying the patch response outputs with learned weights $w_p$ before summation. Finally, we initialise the inducing patch locations $\ell(Z)$ of TICK-GP to random values in $[0, H] \times [0, W]$, and use a Mat\'ern-3/2 kernel with the lengthscale initialised to 3 for the location kernel $k_{\text{loc}}$ from \cref{eq:tick}.

All GP models use a minibatch size of 128 and are trained using the Adam optimiser \citep{kingma2014adam} with a $t^{-1}$ decaying learning rate, starting at $0.01$. The models are run  on a single GeForce GTX 1070 GPU until they converge.

Given that we are dealing with a multiclass classification problem, we use the softmax likelihood with 10 latent GPs. The softmax likelihood is not conjugate to the variational posterior, therefore we evaluate the predictive distribution using Monte Carlo estimates, $\frac{1}{K} \sum\nolimits_k p(y_n \given f^{(k)}(\cdot))$, where $f^{(k)}(\cdot) \sim q(f(\cdot))$. In our experiments we set $K=5$.

\subsection{Deep GPs Comparison (\cref{sec:exp:deep})}
We configure the deep convolutional GP models as identically as possible: each layer uses 384 inducing 5x5 patches (initialised using random patches from the training images), an identity Conv2D mean function for the hidden layers, and a SE kernel for the patch response function. The hidden layers for the L=2 and L=3 models are identical for both the deep Conv-GP and deep TICK-GP, because the translation insensitivity is added to the final layer only. We use a minibatch size of 32 for MNIST and, 64 for CIFAR. All models are optimised using Adam with an exponentially decaying learning rate, starting at 0.01 and decreasing every 50,000 optimisation steps by a factor of 4. We run all models for 300,000 iterations.

\begin{figure}[!ht]
\centering
\includegraphics[width=0.9\linewidth]{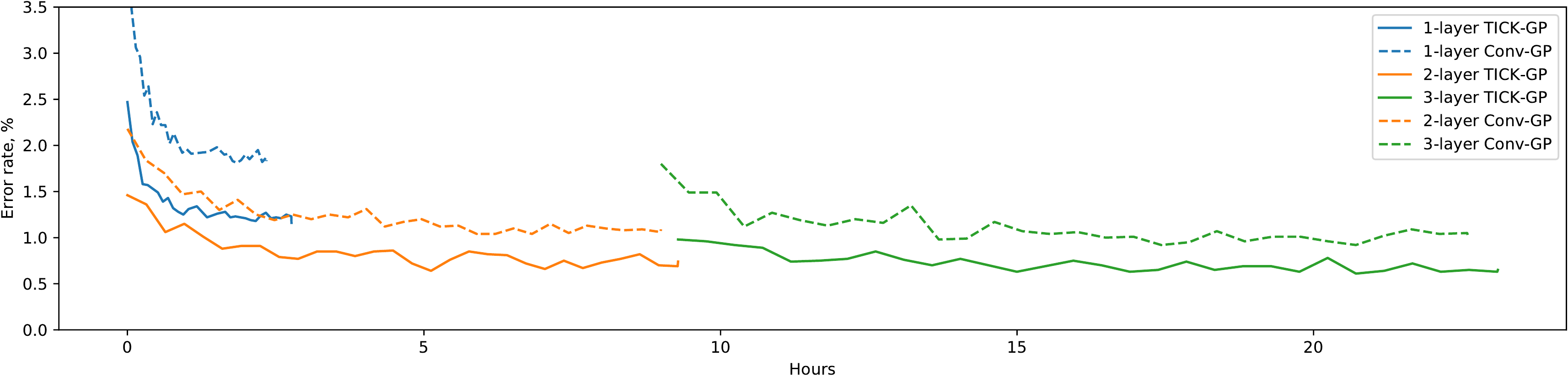}
\caption{Deep convolutional GP error rate traces in function of optimisation time on the MNIST dataset. We plot TICK (solid) and Conv-GP (dashed) models, with one (blue), two (orange), and three (green) layers. All models ran for 300,000 iterations. The three-layered models are initialised with the trained values of the two-layered model.}
\label{fig:appendix:traces-dgp}
\end{figure}

For the initialisation of the hidden layers' variational parameters, we follow \citet{salimbeni2017doubly} and set $\vm = \bm{0}$ and $\MS = \Eye \cdot 10^{-6}$.  The zero mean and small covariance turn off the non-linear GP behaviour of the first layers, making them practically deterministic and completely determined by their identity mean function. In the final layer we set $\vm = \bm{0}$ and $\MS = \Eye$, as we do for the single-layer models in \cref{sec:exp:mnist}. For the initialisation of the three-layer models we set the first and last layer to the trained values of the two-layered model, as was done in \citet{blomqvist2019deep}. This is why we plot the optimisation curves for the three-layered models after the two-layer models in \cref{fig:appendix:traces-dgp}.

\section{CNN Architectures}
\label{sec:appendix:cnn}

The Convolutional Neural Network (CNN) used in the classification experiments consists of two convolutional layers. The convolutional layers are configured to have 32 and 64 kernels respectively, a kernel size of 5x5, and a stride of 1. Both convolutional layers are followed by max pooling with strides and size equal to 2. The output of the second max pooling layer of size 1024 is fed into a fully connected layer with ReLU activation, the result of which is passed through a dropout layer with rate 0.5. The final fully connected layer has 10 units with softmax non-linearity. We initialised the convolutional and fully-connected weights by a truncated normal with standard deviation equal to 0.1. The bias weights were initialised to 0.1 constant.  The CNN is trained using the Adam optimiser described in \cite{kingma2014adam} with a constant learning rate of $0.0001$. We followed the architecture used in \citet{cnn2017tf}.

\section{Predictive Probabilities Of Misclassified MNIST Images}
\label{sec:appendix:classification-images}

\begin{figure}[ht]
\centering
\includegraphics[width=\columnwidth]{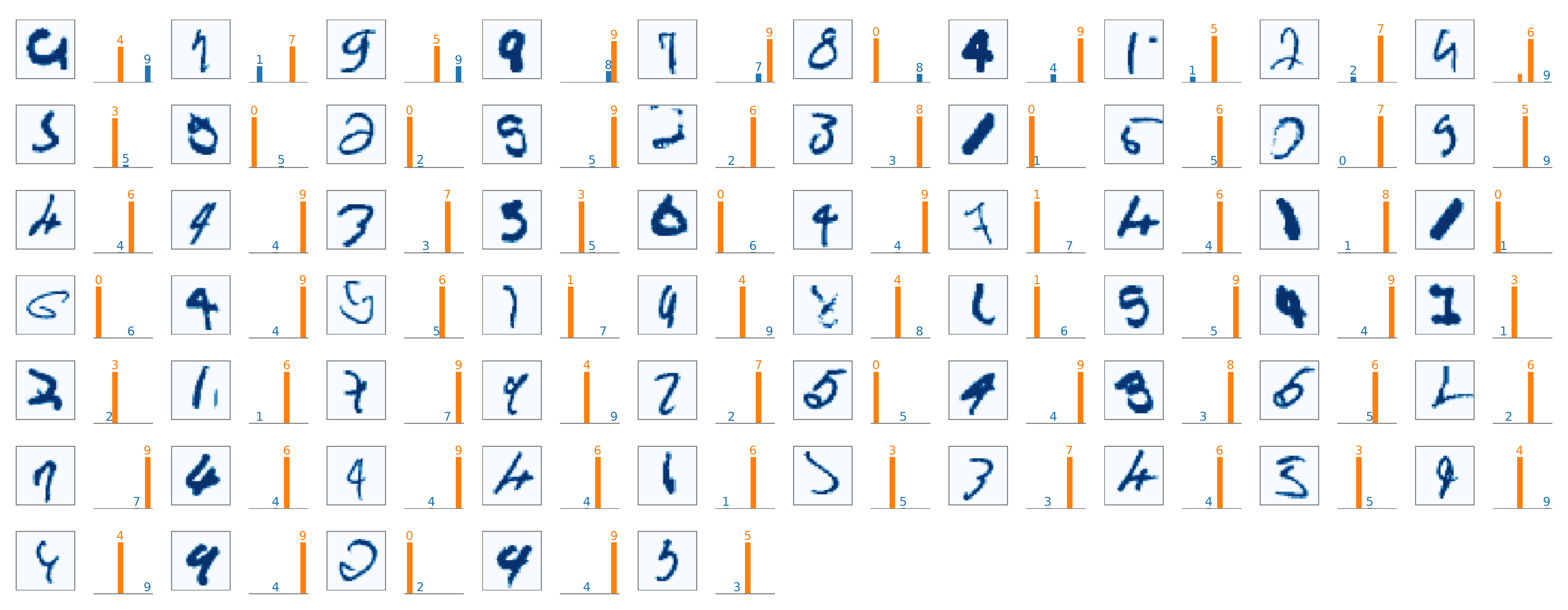}
\caption{CNN model's prediction probabilities for misclassified MNIST images.}
\label{fig:missclassifications1-appendix}
\end{figure}

\begin{figure}[ht]
\centering
\includegraphics[width=\columnwidth]{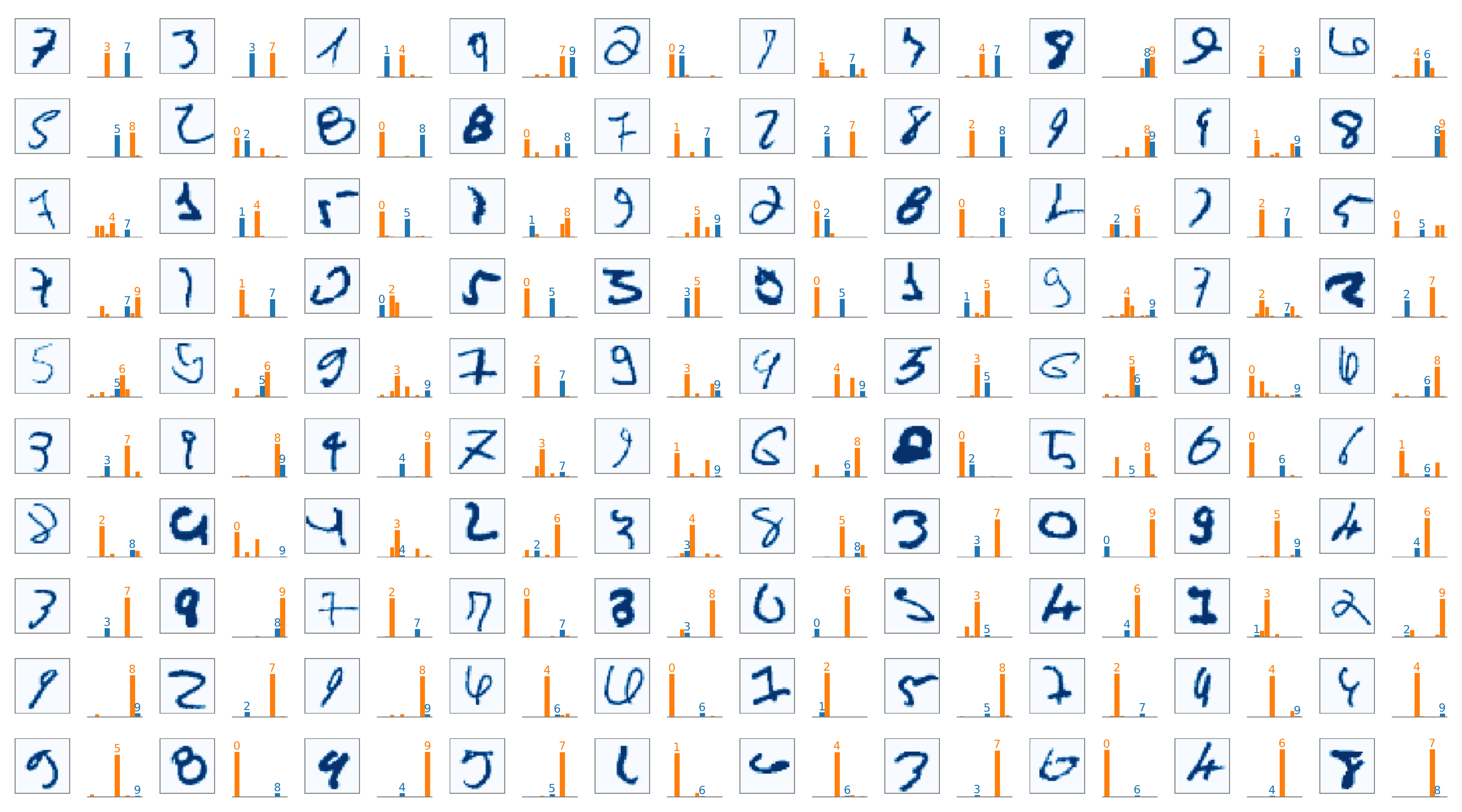}
\caption{Conv-GP model's prediction probabilities for misclassified MNIST images.}
\label{fig:missclassifications2-appendix}
\end{figure}

\begin{figure}[ht]
\centering
\includegraphics[width=\columnwidth]{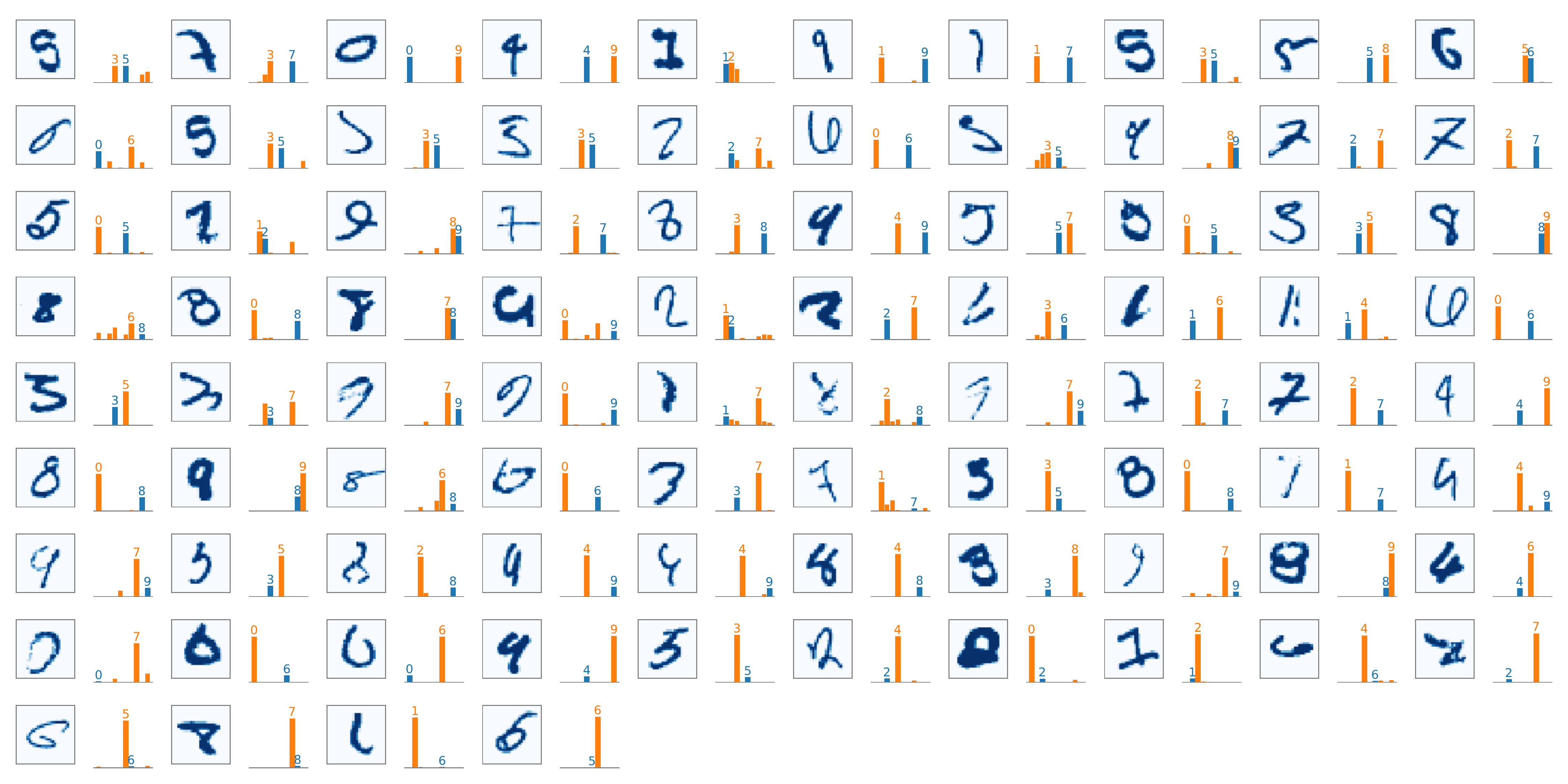}
\caption{TICK-GP model's prediction probabilities for misclassified MNIST images.}
\label{fig:missclassifications3-appendix}
\end{figure}

\section{Out-Of-Distribution Test}
\label{sec:appendix:ood}

In this experiment we test the generalisation capacity of the models presented in \cref{sec:exp:mnist}. In particular, we are interested in studying their behaviour when a distribution shift occurs on the test set. This is an important application, because most machine learning models will eventually be used in domains broader than their training dataset. It is therefore crucial that the models can detect this change of environment, and adjust their uncertainty levels so that appropriate actions can be taken.

The models in \cref{tab:semeion-classification} are trained on MNIST, but the reported metrics, error rate, and NLL are calculated for the Semeion digit dataset. The Semeion dataset \citep{semeion2008} has 1,593 images of 16x16 pixels size. To be able to re-use MNIST trained models, we pad the Semeion images with zero pixels to match the MNIST size. The table shows that TICK-GP outperforms the CNN, and to a lesser extent the Conv-GP, in terms of NLL, and performs comparably to a CNN in terms of accuracy. In \cref{fig:mnist-posterior-probs} we show the predictive probability for the models for a few randomly selected \emph{misclassified} images. The image clearly illustrates the fact that the CNN is making wrong predictions with a very high certainty, explaining the low NLL values.

\begin{table}[ht]
\centering
\caption{Results of Out-Of-Distribution test set experiment. The models are trained on MNIST digits and tested on the different Semeion digit dataset (lower is better). \label{tab:semeion-classification}}
\vspace{.3cm}
\scalebox{0.9}{\begin{tabular}{llll}
\toprule
 metric & Conv-GP & CNN & TICK-GP \\
\midrule
top-$1$ error       & $36.72$       & \bm{$14.44$}     & $16.26$ \\
top-$2$ error       & $16.63$      & \bm{$5.27$}     & $5.71$ \\
top-$3$ error       & $9.10$      & $1.95$     & \bm{$1.76$} \\
\midrule
NLL full test set   & $1.027$      & $2.115$     & \bm{$0.474$} \\
NLL misclassified   & $2.221$      & $14.614$     & \bm{$1.941$} \\
\bottomrule
\end{tabular}}
\end{table}

\begin{figure}[!ht]
\centering
\includegraphics[width=0.48\linewidth]{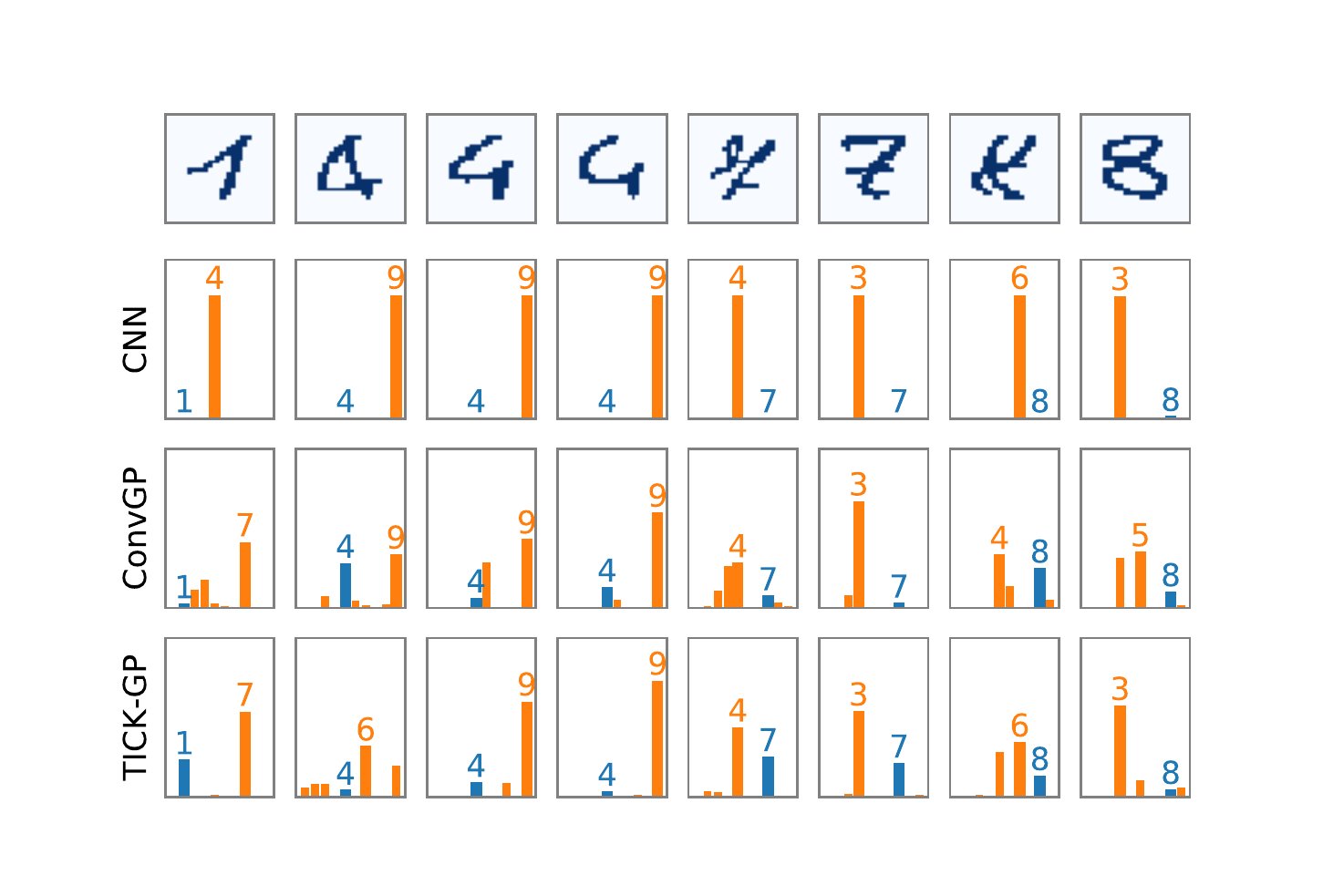}
\caption{Prediction probabilities for eight randomly selected misclassified images (top row) form the Semeion dataset. The bars show the probabilities for each of the classes, 0 to 9. The largest orange bar is the class with highest probability and is thus used as a prediction from the model; the blue bar is the true class label.}
\label{fig:semeion-posterior-probs}
\end{figure}

\begin{figure}[!ht]
\centering
\includegraphics[width=0.7\linewidth]{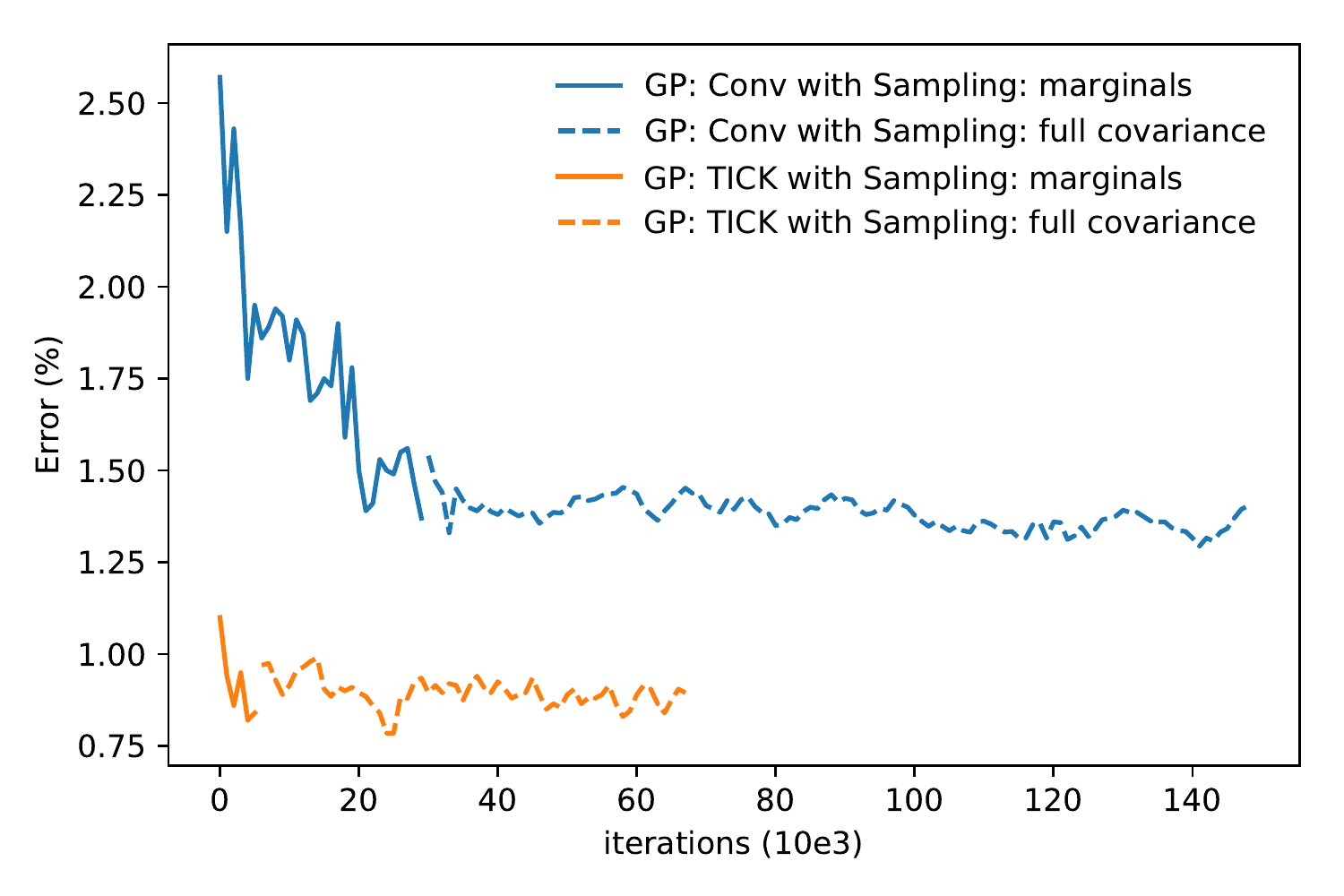}
\caption{In this experiment we run two 2-layered deep convolutional models: one using the TICK kernel and another using the original convolutional kernel. We first optimise the models by sampling from only the marginals of the hidden layer's posterior GPs $q(f_\ell(\cdot))$, and then switch to sampling from the full covariance. We see the performance of the models slightly improving, but not enough to justify the added computational complexity. These are costly experiments, which took roughly 10 days to run.}
\label{fig:sampling-full-cov}
\end{figure}

\end{appendices}

\end{document}